\PassOptionsToPackage{usenames,dvipsnames,table}{xcolor}
\documentclass[11pt,a4paper]{article}
\usepackage{naaclhlt2018}
\aclfinalcopy
\usepackage{times}
\usepackage[utf8]{inputenc} % allow utf-8 input
\usepackage[T1]{fontenc}    % use 8-bit T1 fonts
\usepackage{url}            % simple URL typesetting
\usepackage{booktabs}       % professional-quality tables
\usepackage{amsfonts}       % blackboard math symbols

\usepackage{nicefrac}       % compact symbols for 1/2, etc.
\usepackage{microtype}      % microtypography
\usepackage{latexsym}
\usepackage{amssymb}
\usepackage{amsmath}
\usepackage{multirow}
\usepackage{rotating}
\usepackage{subfig}
\usepackage{array}
\usepackage{enumitem}% http://ctan.org/pkg/enumitem
\usepackage{makecell}
\usepackage{xcolor}
\usepackage[toc,page]{appendix}
\usepackage{xspace}
\usepackage{mdwlist}
\usepackage{graphics}
\usepackage{color}
\usepackage{epsfig}
\usepackage{alltt}
\usepackage{moreverb}
\usepackage{fancyvrb}
\usepackage{enumerate}
\usepackage{colortbl}           % for color table cells
\usepackage{relsize}
\usepackage{txfonts}
\usepackage[labelfont=bf]{caption}
\usepackage{float}
\usepackage{etoolbox}
\usepackage{arydshln}
\usepackage{setspace}
\usepackage{mathtools}
\usepackage{bm}
\usepackage{soul}
\usepackage[draft]{pgf}
\usepackage[normalem]{ulem}

\usepackage{tabularx}
\definecolor{vlgray}{gray}{0.95}

\newcolumntype{L}[1]{>{\raggedright\let\newline\\\arraybackslash\hspace{0pt}}m{#1}}
\newcolumntype{C}[1]{>{\centering\let\newline\\\arraybackslash\hspace{0pt}}m{#1}}
\newcolumntype{R}[1]{>{\raggedleft\let\newline\\\arraybackslash\hspace{0pt}}m{#1}}

\usepackage{lscape}
 \newcolumntype{b}{>{\hsize=2.3\hsize}X}
\newcolumntype{s}{>{\hsize=.45\hsize}X}
\newcolumntype{m}{>{\hsize=.9\hsize}X}

\newtoggle{draft}
\toggletrue{draft}
\iftoggle{draft}{
\newcommand{\roy}[1]{\textcolor{BurntOrange}{[#1 \textsc{--Roy}]}}
\newcommand{\wacomment}[1]{\textcolor{Blue}{[#1 \textsc{--Waleed}]}}
\newcommand{\bhavana}[1]{\textcolor{Green}{[#1 \textsc{--Bhavana}]}}
\newcommand{\dk}[1]{\textcolor{Maroon}{[#1 \textsc{--DK}]}}
\newcommand{\kylel}[1]{\textcolor{Bittersweet}{[#1 \textsc{--Kyle}]}}

}{
\newcommand{\roy}[1]{}
\newcommand{\wacomment}[1]{}
\newcommand{\bhavana}[1]{}
\newcommand{\dk}[1]{}
\newcommand{\kylel}[1]{}
}

\newcommand{\camready}[1]{}
\newenvironment{itemizesquish}{\begin{list}{\labelitemi}{\setlength{\itemsep}{0em}\setlength{\labelwidth}{0.5em}\setlength{\leftmargin}{\labelwidth}\addtolength{\leftmargin}{\labelsep}}}{\end{list}}

\newcommand{\com}[1]{}
\makeatletter
\g@addto@macro\normalsize{%
  \setlength\abovedisplayskip{1pt}
  \setlength\belowdisplayskip{1pt}
  \setlength\abovedisplayshortskip{1pt}
 \setlength\belowdisplayshortskip{1pt}
}
\renewcommand{\paragraph}{%
  \@startsection{paragraph}{4}%
  {\z@}{0.15ex \@plus 0.5ex \@minus .2ex}{-1em}%
  {\normalfont\normalsize\bfseries}%
}
\makeatother

\title{Construction of the Literature Graph in Semantic Scholar}

\author{
\parbox{\linewidth}{\centering
Waleed Ammar, Dirk Groeneveld, Chandra Bhagavatula, Iz Beltagy, Miles Crawford,
Doug Downey,$^\dagger$
Jason Dunkelberger, 
Ahmed Elgohary, 
Sergey Feldman,
Vu Ha,\\
Rodney Kinney, 
Sebastian Kohlmeier, 
Kyle Lo,
Tyler Murray, 
Hsu-Han Ooi, \\
Matthew Peters, 
Joanna Power, 
Sam Skjonsberg, 
Lucy Lu Wang, 
Chris Wilhelm,\\
Zheng Yuan,$^\dagger$
Madeleine van Zuylen,
and Oren Etzioni} \\
\texttt{waleeda@allenai.org} \\ 
\\ Allen Institute for Artificial Intelligence, Seattle WA 98103, USA
\\ $^\dagger$Northwestern University, Evanston IL 60208, USA
}
\date{}

\hypersetup{draft}
\begin{document}
\maketitle

\begin{abstract}
We describe a deployed scalable system for organizing published scientific literature into a heterogeneous graph to facilitate algorithmic manipulation and discovery.
The resulting literature graph consists of more than 280M nodes, representing papers, authors, entities and various interactions between them (e.g., authorships, citations, entity mentions).
We reduce literature graph construction into familiar NLP tasks (e.g., entity extraction and linking), point out research challenges due to differences from standard formulations of these tasks, and report empirical results for each task.
The methods described in this paper are used to enable semantic features in \url{www.semanticscholar.org}.
\end{abstract}

\section{Introduction}
The goal of this work is to facilitate algorithmic discovery in the scientific literature.
Despite notable advances in scientific search engines, data mining and digital libraries \citep[e.g.,][]{wu:14}, researchers remain unable to answer simple questions such as:
\begin{itemizesquish}
\item What is the percentage of female subjects in depression clinical trials?
\item Which of my co-authors published one or more papers on coreference resolution?
\item Which papers discuss the effects of Ranibizumab on the Retina?
%\item How many times was \newcite{lafferty:01} cited at ACL vs. CVPR?
\end{itemizesquish}

\begin{figure}[t]	
\centering
{
\includegraphics[width=0.8\linewidth]{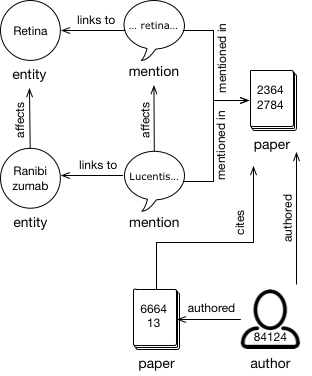}
}
\caption{Part of the literature graph. \label{fig:graphlet}}
\end{figure}

In this paper, we focus on the problem of extracting structured data from scientific documents, which can later be used in natural language interfaces \cite[e.g.,][]{iyer:17} or to improve ranking of results in academic search \cite[e.g.,][]{xiong:17}.
We describe methods used in a scalable deployed production system for extracting structured information from scientific documents into \emph{the literature graph} (see Fig. \ref{fig:graphlet}).
The literature graph is a directed property graph which summarizes key information in the literature and can be used to answer the queries mentioned earlier as well as more complex queries.
For example, in order to compute the Erd\H{o}s number of an author X, the graph can be queried to find the number of nodes on the shortest undirected path between author X and Paul Erd\H{o}s such that all edges on the path are labeled ``authored''.

We reduce literature graph construction into familiar NLP tasks such as sequence labeling, entity linking and relation extraction, and address some of the impractical assumptions commonly made in the standard formulations of these tasks.
For example, most research on named entity recognition tasks report results on large labeled datasets such as CoNLL-2003 and ACE-2005 \cite[e.g.,][]{lample:16}, and assume that entity types in the test set match those labeled in the training set \cite[including work on domain adaptation, e.g.,][]{daume:07}.
These assumptions, while useful for developing and benchmarking new methods, are unrealistic for many domains and applications.
The paper also serves as an overview of the approach we adopt at \url{www.semanticscholar.org} in a step towards more intelligent academic search engines \cite{etzioni:11}.

In the next section, we start by describing our symbolic representation of the literature. 
Then, we discuss how we extract metadata associated with a paper such as authors and references, 
%in \S\ref{sec:science_parse}, 
then how we extract the entities mentioned in paper text.
%in \S\ref{sec:entities}
Before we conclude,
%in \S\ref{sec:others}, 
we briefly describe other research challenges we are actively working on in order to improve the quality of the literature graph. 

%The resulting literature graph consists of more than 280 million nodes, representing papers, authors, entities and various interactions between them (e.g., authorships, citations, entity mentions).
%In the following section, we discuss the structure of the literature graph.

\section{Structure of The Literature Graph}\label{sec:structure}
The literature graph is a \textit{property graph} with directed edges.
Unlike Resource Description Framework (RDF) graphs, nodes and edges in property graphs have an internal structure which is more suitable for representing complex data types such as papers and entities.
In this section, we describe the attributes associated with nodes and edges of different types in the literature graph.

\subsection{Node Types}
% paper nodes => partnerships + science-parse.
\paragraph{Papers.}
We obtain metadata and PDF files of papers via partnerships with publishers (e.g., Springer, Nature), catalogs (e.g., DBLP, MEDLINE), pre-publishing services (e.g., arXiv, bioRxive), as well as web-crawling. 
Paper nodes are associated with a set of attributes such as `title', `abstract', `full text', `venues' and `publication year'.
While some of the paper sources provide these attributes as metadata, it is often necessary to extract them from the paper PDF (details in \S\ref{sec:science_parse}).
We deterministically remove duplicate papers based on string similarity of their metadata, resulting in 37M unique paper nodes.
%\kylel{similarity of metadata or attributes?} \wacomment{TODO: I'm not clear on the details here actually. We'll need to clarify this with someone from the offline team before camera ready. }
Papers in the literature graph cover a variety of scientific disciplines, including computer science, molecular biology, microbiology and neuroscience.

% author nodes => partnerships + science-parse + author disambiguation.
\paragraph{Authors.}
Each node of this type represents a unique author, with attributes such as `first name' and `last name'.
The literature graph has 12M nodes of this type.
%We describe how we identify unique authors in \S\ref{sec:author_disambiguation}.

% entity nodes => KB imports + knowledge discovery.
\paragraph{Entities.}
Each node of this type represents a unique scientific concept discussed in the literature, with attributes such as `canonical name', `aliases' and `description'.
Our literature graph has 0.4M nodes of this type.
We describe how we populate entity nodes in \S\ref{sec:kbs}.

% mention nodes => entity extraction + linking.
\paragraph{Entity mentions.}
Each node of this type represents a textual reference of an entity in one of the papers, with attributes such as `mention text', `context', and `confidence'. 
We describe how we populate the 237M mentions in the literature graph in \S\ref{sec:entities_approaches}.
%\kylel{The intro said 50M nodes, but this is easily 200M+?}\wacomment{fixed}

\subsection{Edge Types}
% citation edges => science-parse + partnerships.
\paragraph{Citations.}
We instantiate a directed citation edge from paper nodes $p_1 \longrightarrow p_2$ for each $p_2$ referenced in $p_1$.
Citation edges have attributes such as `from paper id', `to paper id' and `contexts' (the textual contexts where $p_2$ is referenced in $p_1$).
While some of the paper sources provide these attributes as metadata, it is often necessary to extract them from the paper PDF as detailed in \S\ref{sec:science_parse}.
%\kylel{rephrase/shorten to: "extract them from the paper PDF."} \wacomment{done}
%The literature graph has 334M edges of this type. 
%\wacomment{TODO: check paper coverage with Hsu-Han/Field.}

% authorship edges => science-parse.
\paragraph{Authorship.}
We instantiate a directed authorship edge between an author node and a paper node $a \longrightarrow p$ for each author of that paper.
%Due to the lack of a global ID for authors, it is non-trivial to identify which author nodes should be connected to each paper (see \S\ref{sec:author_disambiguation} for details).
%In total, \wacomment{TODO: check number of edges with Hsu-Han/Field.}

\paragraph{Entity linking edges.}
We instantiate a directed edge from an extracted entity mention node to the entity it refers to. 
%Note that this edge is only instantiated when the paper analyzers described in \S\ref{sec:entities} identify the entity with enough confidence.
%\kylel{a bit confusing this is same name as previous node section called Entity mentions.  Maybe "Entity-mention relations"}\wacomment{fixed.}

% mention-mention edges => relation extraction.
\paragraph{Mention--mention relations.}
We instantiate a directed edge between a pair of mentions in the same sentential context if the textual relation extraction model predicts one of a predefined list of relation types between them in a sentential context.\footnote{Due to space constraints, we opted not to discuss our relation extraction models in this draft.}
We encode a symmetric relation between $m_1$ and $m_2$ as two directed edges $m_1 \longrightarrow m_2$ and $m_2 \longrightarrow m_1$.
%See \ref{sec:knowledge_discovery} for more details on relation extraction.

% entity-mention edges => entity extraction and linking.

%\lwcomment{Maybe swap the ordering of the two above edge types? The first references the second.}\wacomment{done}

% entity-entity edges => KB imports + KB completion.
\paragraph{Entity--entity relations.}\qquad
While mention--mention edges represent relations between mentions in a particular context, entity--entity edges represent relations between abstract entities.
These relations may be imported from an existing knowledge base  (KB) or inferred from other edges in the graph.
%(see details in \S\ref{sec:knowledge_discovery}).

\section{Extracting Metadata}\label{sec:science_parse}

In the previous section, we described the overall structure of the literature graph. 
Next, we discuss how we populate paper nodes, author nodes, authorship edges, and citation edges.

% define problem: given pdf file, label author names and references.
%To construct paper nodes, author nodes, and citation edges, we rely primarily on metadata provided by publishers. Some publishers don't provide metadata, and of those who do, the data is often incomplete. 
Although some publishers provide sufficient metadata about their papers, many papers are provided with incomplete metadata. 
Also, papers obtained via web-crawling are not associated with any metadata.
To fill in this gap, we built the \textsc{ScienceParse} system to predict structured data from the raw PDFs using recurrent neural networks (RNNs).\footnote{The \textsc{ScienceParse} libraries can be found at \url{http://allenai.org/software/}.}
For each paper, the system extracts the paper title, list of authors, and list of references; each reference consists of a title, a list of authors, a venue, and a year.

%We split each PDF into its individual pages, tokenize each page, and use an LSTM to label each token as one of seven classes: `none', `title', `author', `bib title', `bib author', `bib year', and `bib venue'. 
%We then apply some simple heuristics to construct paper nodes, author nodes and citation edges from the labeled tokens.

\paragraph{Preparing the input layer.}
We split each PDF into individual pages, and feed each page to Apache's PDFBox library\footnote{\url{https://pdfbox.apache.org}} to convert it into a sequence of tokens, where each token has features, e.g., `text', `font size', `space width', `position on the page'.
%While tokenizing a PDF is not trivial, the heuristics built into PDFBox work fairly well, and only have to be augmented in a few corner cases. 

We normalize the token-level features before feeding them as inputs to the model.
For each of the `font size' and `space width' features, we compute three normalized values (with respect to current page, current document, and the whole training corpus), each value ranging between -0.5 to +0.5.
The token's `position on the page' is given in XY coordinate points. We scale the values linearly to range from $(-0.5,-0.5)$ at the top-left corner of the page to $(0.5,0.5)$ at the bottom-right corner.
%\kylel{Not clear to me if each point tuple is mapped to a single scalar, or if both coordinates are scaled}\wacomment{TODO: clarify with Dirk.}

In order to capture case information, we add seven numeric features to the input representation of each token: whether the first/second letter is uppercase/lowercase, the fraction of uppercase/lowercase letters and the fraction of digits.
%The uppercase features are not completely complimentary to the lowercase ones. 
%Characters can be neither uppercase nor lowercase, for example for digits, mathematical notation, or non-Latin alphabets.
%\kylel{Based on this paragraph, inclined to move this to Preprocessing section.  But also, seems like Preprocessing section is closer to Feature engineering, especially for `font' and `page number' features} \wacomment{+1, moved}

To help the model make correct predictions for metadata which tend to appear at the beginning (e.g., titles and authors) or at the end of papers (e.g., references), we provide the current page number as two discrete variables (relative to the beginning and end of the PDF file) with values 0, 1 and 2+.
These features are repeated for each token on the same page.
%For the `page number', we compute both the number of pages from the front and the the number of pages from the back. This helps the network distinguish between `title' and `author', usually at the beginning of the PDF, and the `bibliography' classes, usually at the end. 
%Both numbers are clipped \kylel{TODO: verify this is a clipping and not some linear scaling} to range from 0 to 2.
% \kylel{@waleed:  this paragraph seems like its not going through a neural embedding, but more a normalization/transformation of a token feature that belongs in previous section?} %\wacomment{+1, dirk may disagree though.}

For the $k$-th token in the sequence, we compute the input representation $\mathbf{i}_k$ by concatenating the numeric features, an embedding of the `font size', and the word embedding of the lowercased token.
Word embeddings are initialized with GloVe \cite{pennington:14}.

\paragraph{Model.}
%The embedding layer loses all information about capitalization, which is an important feature for detecting `titles' and `authors'. 
%\kylel{Not sure if "we could've done this, but.." is necessary; seems distracting -->} \wacomment{+1, commented out for now}
%To remedy this, we could make capitalization part of the embedding (i.e., don't normalize the capitalization before embedding the token), but that would result in a sparser representation, less training data per token, and more independent weights to train. 
% describe model.
%All of these pre-processed features are concatenated, and passed through one fully-connected layer to produce a final vector representation for each token.
%\paragraph{LSTM and output}
The input token representations are passed through one fully-connected layer and then fed into a two-layer bidirectional LSTM \cite[Long Short-Term Memory,][]{hochreiter:97}, i.e.,

\begin{align*}
\mathbf{g}_k^\rightarrow &= \text{LSTM}(\mathbf{Wi}_k, \mathbf{g}_{k-1}^\rightarrow), \mathbf{g}_k = [\mathbf{g}_k^\rightarrow; \mathbf{g}_k^\leftarrow], \nonumber \\
\mathbf{h}_k^\rightarrow &= \text{LSTM}(\mathbf{g}_k, \mathbf{h}_{k-1}^\rightarrow), \mathbf{h}_k = [\mathbf{h}_k^\rightarrow; \mathbf{g}_k^\leftarrow] \nonumber \\
\end{align*}
where $W$ is a weight matrix, $\mathbf{g}_k^\leftarrow$ and $\mathbf{h}_k^\leftarrow$ are defined similarly to $\mathbf{g}_k^\rightarrow$ and $\mathbf{h}_k^\rightarrow$ but process token sequences in the opposite direction.

Following \newcite{collobert:11}, we feed the output of the second layer $\mathbf{h}_k$ into a dense layer to predict unnormalized label weights for each token and learn label bigram feature weights (often described as a conditional random field layer when used in neural architectures) to account for dependencies between labels.

\paragraph{Training.}
The \textsc{ScienceParse} system is trained on a snapshot of the data at PubMed Central. It consists of 1.4M PDFs and their associated metadata, which specify the correct titles, authors, and bibliographies. We use a heuristic labeling process that finds the strings from the metadata in the tokenized PDFs to produce labeled tokens. This labeling process succeeds for 76\% of the documents. The remaining documents are not used in the training process.
During training, we only use pages which have at least one token with a label that is not ``none''. 
%\wacomment{TODO: not clear why this labeling process does not succeed 100% of the time.}
% TODO: Doug or Zheng to confirm that the 10% number is still correct with bibs.

%\paragraph{Training process}
%\wacomment{had to comment out this paragraph because it has too much hand waving.}
%Most of a paper's tokens are in the body of the paper, and receive a label of `none'. This hurts training,
%\wacomment{TODO: clarify whether it hurts runtime or performance?}
%since there is little information in the `none' tokens. 
%We use a variety of techniques to address this problem. 
%In general, we only train on pages that have at least one token with a label that's not `none'. 
%That means the model sees a different distribution of tokens during training and test, but in practice, it improves training times with no adverse effect on quality. 

%\wacomment{TODO: refactor for the camera ready. describe as curriculum learning.}
%Beyond that, we employ a special training regimen to make sure the model sees enough title tokens. 
%As a first pass, we train on only the first three pages of every document. That way we see a lot of titles, but not a lot of bibliographies. We train this way until the model stops improving. In a second pass, we train on the first 50 pages of the documents (i.e., all pages for over 99\% of the documents). This pass learns to correctly classify the bibliography tokens, without compromising too much quality on titles and authors.

\paragraph{Decoding.}
%The labeled tokens need a little bit of post-processing to produce the final structured output. 
At test time, we use Viterbi decoding to find the most likely global sequence, with no further constraints.
To get the title, we use the longest continuous sequence of tokens with the ``title'' label. 
Since there can be multiple authors, we use all continuous sequences of tokens with the ``author'' label as authors, but require that all authors of a paper are mentioned on the same page.
If the author labels are predicted in multiple pages, we use the one with the largest number of authors.

%Titles in the bibliography section often contain words that are hyphenated across two lines, resulting in three labeled tokens, the first part of the word, the hyphen, and the second part of the word. In post-processing, we detect and fix this condition to produce clean output.

\paragraph{Results.}
\begin{table}
\centering
\begin{tabular}{@{}r|ccc@{}}
\toprule
Field                & Precision & Recall & F1    \\ \midrule
title                & 85.5     & 85.5  & 85.5 \\
authors              & 92.1     & 92.1  & 92.1 \\
bibliography titles  & 89.3     & 89.4  & 89.3 \\
bibliography authors & 97.1     & 97.0  & 97.0 \\
bibliography venues  & 91.7     & 89.7  & 90.7 \\
bibliography years   & 98.0     & 98.0  & 98.0 \\ \bottomrule
\end{tabular}
\caption{Results of the \textsc{ScienceParse} system.}
\label{tab:spresults}
\end{table}
We run our final tests on a held-out set from PubMed Central, consisting of about 54K documents. The results are detailed in Table \ref{tab:spresults}. 
We use a conservative evaluation where an instance is correct if it exactly matches the gold annotation, with no credit for partial matching. 

To give an example for the type of errors our model makes, consider the paper \cite{wang:13} titled ``Clinical review: Efficacy of antimicrobial-impregnated catheters in external ventricular drainage - a systematic review and meta-analysis.'' 
The title we extract for this paper omits the first part ``Clinical review:''.
This is likely to be a result of the pattern ``Foo: Bar Baz'' appearing in many training examples with only ``Bar Baz'' labeled as the title.

\section{Entity Extraction and Linking}\label{sec:entities}

In the previous section, we described how we populate the backbone of the literature graph, i.e., paper nodes, author nodes and citation edges.
Next, we discuss how we populate mentions and entities in the literature graph using entity extraction and linking on the paper text. 
In order to focus on more salient entities in a given paper, we only use the title and abstract.

\subsection{Approaches}\label{sec:entities_approaches}
We experiment with three approaches for entity extraction and linking: 
%\kylel{The definition of a "paper analyzer" isnt obvious to me. Maybe sentence preceding this that says "We refer to a set of heuristics or models that extracts mentions from papers and links them to entities in KBs as a paper analyzer."}\wacomment{good call! fixed.}
\vspace{1.5mm}

{\noindent
\textbf{I. Statistical:} uses one or more statistical models for predicting mention spans, then uses another statistical model to link mentions to candidate entities in a KB.
}
\vspace{1.5mm}

{\noindent
\textbf{II. Hybrid:} defines a small number of hand-engineered, deterministic rules for string-based matching of the input text to candidate entities in the KB, then uses a statistical model to disambiguate the mentions.\footnote{We also experimented with a ``pure'' rules-based approach which disambiguates deterministically but the hybrid approach consistently gave better results.}
}
\vspace{1.5mm}

{\noindent
\textbf{III. Off-the-shelf:} uses existing libraries, namely \cite[][TagMe]{ferragina:10}\footnote{The TagMe APIs are described at \url{https://sobigdata.d4science.org/web/tagme/tagme-help}} and \cite[][MetaMap Lite]{demnerfushman:17}\footnote{We use v3.4 (L0) of MetaMap Lite, available at \url{https://metamap.nlm.nih.gov/MetaMapLite.shtml}}, with minimal post-processing to extract and link entities to the KB.
}
\vspace{1.5mm}

% results on end-to-end (rodney's).
% v1.1.3 results: https://docs.google.com/document/d/1DZuKIz1M01i_SUCASd-ezSHhYXe88iqkHg8nxWR0Skk/edit#heading=h.a7gu3tbb2cdc
We evaluate the performance of each approach in two broad scientific areas: computer science (CS) and biomedical research (Bio).
For each unique (paper ID, entity ID) pair predicted by one of the approaches, we ask human annotators to label each mention extracted for this entity in the paper.
We use CrowdFlower to manage human annotations and only include instances where three or more annotators agree on the label.
If one or more of the entity mentions in that paper is judged to be correct, the pair (paper ID, entity ID) counts as one correct instance.
Otherwise, it counts as an incorrect instance.
We report `yield' in lieu of `recall' due to the difficulty of doing a scalable comprehensive annotation.

\begin{table}[t]
\centering
\begin{tabular}{@{}r|cr|cr@{}}
\toprule
Approach & \multicolumn{2}{c}{CS} &  \multicolumn{2}{c}{Bio}  \\ 
         & prec.  &  yield  & prec. & yield \\ \midrule
Statistical  & 98.4 & 712 & 94.4 & 928 \\ 
Hybrid & 91.5 & 1990 & 92.1 & 3126 \\
Off-the-shelf & 97.4 & 873 & 77.5 & 1206 \\
\bottomrule
\end{tabular}
\caption{Document-level evaluation of three approaches in two scientific areas: computer science (CS) and biomedical (Bio). }
\label{tab:analyzers}
\end{table}

Table \ref{tab:analyzers} shows the results based on 500 papers using v1.1.2 of our entity extraction and linking components. 
In both domains, the statistical approach gives the highest precision and the lowest yield.
The hybrid approach consistently gives the highest yield, but sacrifices precision.
The TagMe off-the-shelf library used for the CS domain gives surprisingly good results, with precision within 1 point from the statistical models.
However, the MetaMap Lite off-the-shelf library we used for the biomedical domain suffered a huge loss in precision.
Our error analysis showed that each of the approaches is able to predict entities not predicted by the other approaches so we decided to pool their outputs in our deployed system, which gives significantly higher yield than any individual approach while maintaining reasonably high precision.

\subsection{Entity Extraction Models}
\label{extraction_model}

% define problem.
Given the token sequence $t_1, \ldots, t_N$ in a sentence, we need to identify spans which correspond to entity mentions.
We use the BILOU scheme to encode labels at the token level.
Unlike most formulations of named entity recognition problems (NER), we do not identify the entity type (e.g., protein, drug, chemical, disease) for each mention since the output mentions are further grounded in a KB with further information about the entity (including its type), using an entity linking module.
%\lwcomment{Not sure what entity type is referring to here. Not mentioned previously. Maybe just remove this sentence?}\wacomment{clarified with examples}

%\wacomment{TODO: discuss how this differs from standard NER problems: lack of labeled data, entity types, high-precision.}

% describe model.
\paragraph{Model.}
%\wacomment{consider adding a figure.}
First, we construct the token embedding $\mathbf{x}_k = [\mathbf{c}_k; \mathbf{w}_k]$ for each token $t_k$ in the input sequence, where $\mathbf{c}_k$ is a character-based representation computed using a convolutional neural network (CNN) with filter of size 3 characters, and $\mathbf{w}_k$ are learned word embeddings initialized with the GloVe embeddings \cite{pennington:14}.

We also compute context-sensitive word embeddings, denoted as $\mathbf{lm}_k = [\mathbf{lm}_k^{\rightarrow};\mathbf{lm}_k^{\leftarrow}]$, by concatenating the projected outputs of forward and backward recurrent neural network language models (RNN-LM) at position $k$.
The language model (LM) for each direction is trained independently and consists of a single layer long short-term memory (LSTM) network followed by a linear project layer. 
While training the LM 
%\kylel{we know that LM means "language model"?} \wacomment{fixed.}
parameters, $\mathbf{lm}^{\rightarrow}_k$ is used to predict $t_{k+1}$ and $\mathbf{lm}^{\leftarrow}_k$ is used to predict $t_{k-1}$.
We fix the LM parameters during training of the entity extraction model.
See \newcite{peters:17} and \newcite{ammar:17} for more details.

Given the $\mathbf{x}_k$ and $\mathbf{lm}_k$ embeddings for each token $k \in \{1, \ldots, N \}$, we use a two-layer bidirectional LSTM to encode the sequence with $\mathbf{x}_k$ and $\mathbf{lm}_k$ feeding into the first and second layer, respectively. That is,
% $\mathbf{g}_k^\rightarrow = \text{LSTM}(\mathbf{x}_k, \mathbf{g}_{k-1}^\rightarrow)$, $\mathbf{g}_k = [\mathbf{g}_k^\rightarrow ; \mathbf{g}_k^\leftarrow]$,
% $\mathbf{h}_k^\rightarrow = \text{LSTM}([\mathbf{g}_k ; \mathbf{lm}_k], \mathbf{h}_{k-1}^\rightarrow)$,
% $\mathbf{h}_k = [\mathbf{h}_k^\rightarrow; \mathbf{h}_k^\leftarrow]$,
\begin{align*} 
\mathbf{g}_k^\rightarrow &= \text{LSTM}(\mathbf{x}_k, \mathbf{g}_{k-1}^\rightarrow), 
\mathbf{g}_k = [\mathbf{g}_k^\rightarrow ; \mathbf{g}_k^\leftarrow], \nonumber \\
\mathbf{h}_k^\rightarrow &= \text{LSTM}([\mathbf{g}_k ; \mathbf{lm}_k], \mathbf{h}_{k-1}^\rightarrow),
\mathbf{h}_k = [\mathbf{h}_k^\rightarrow; \mathbf{h}_k^\leftarrow], \nonumber 
\end{align*}
where $\mathbf{g}_k^\leftarrow$ and $\mathbf{h}_k^\leftarrow$ are defined similarly to $\mathbf{g}_k^\rightarrow$ and $\mathbf{h}_k^\rightarrow$ but process token sequences in the opposite direction.

Similar to the model described in \S\ref{sec:science_parse}, we feed the output of the second LSTM into a dense layer to predict unnormalized label weights for each token and learn label bigram feature weights to account for dependencies between labels. 

% report results from the ScienceIE shared task and from the ACL 2017 paper.
\paragraph{Results.}
We use the standard data splits of the SemEval-2017 Task 10 on entity (and relation) extraction from scientific papers \cite{augenstein:17}.
Table \ref{tab:sciencie_entities} compares three variants of our entity extraction model.
The first line omits the LM embeddings $\mathbf{lm}_k$, while the second line is the full model (including LM embeddings) showing a large improvement of 4.2 F1 points.
The third line shows that creating an ensemble of 15 models further improves the results by 1.1 F1 points.

\begin{table}[t]
\centering
\begin{tabular}{@{}r|c@{}}
\toprule
Description & F1    \\ \midrule
Without LM   & 49.9 \\ 
With LM & 54.1 \\
Avg. of 15 models with LM & 55.2 \\
\bottomrule
\end{tabular}
\caption{Results of the entity extraction model on the development set of SemEval-2017 task 10.}
\label{tab:sciencie_entities}
\end{table}

\paragraph{Model instances.}
In the deployed system, we use three instances of the entity extraction model with a similar architecture, but trained on different datasets.
Two instances are trained on the BC5CDR \cite{li:16} and the CHEMDNER datasets \cite{krallinger:15} to extract key entity mentions in the biomedical domain such as diseases, drugs and chemical compounds.
The third instance is trained on mention labels induced from Wikipedia articles in the computer science domain. %\wacomment{TODO: elaborate on how we induce labels from Wikipedia articles.}
The output of all model instances are pooled together and combined with the rule-based entity extraction module, then fed into the entity linking model (described below).

\subsection{Knowledge Bases}\label{sec:kbs}
% describe where our entities and relations come from.
In this section, we describe the construction of entity nodes and entity-entity edges. 
Unlike other knowledge extraction systems such as the Never-Ending Language Learner (NELL)\footnote{\url{http://rtw.ml.cmu.edu/rtw/}} and OpenIE 4,\footnote{\url{https://github.com/allenai/openie-standalone}} we use existing knowledge bases (KBs) of entities to reduce the burden of identifying coherent concepts. 
Grounding the entity mentions in a manually-curated KB also increases user confidence in automated predictions.
We use two KBs:

{\noindent
\textbf{UMLS:} The UMLS metathesaurus integrates information about concepts in specialized ontologies in several biomedical domains, and is funded by the U.S. National Library of Medicine.
}

{\noindent
\textbf{DBpedia:} DBpedia provides access to structured information in Wikipedia. 
Rather than including all Wikipedia pages, we used a short list of Wikipedia categories about CS and included all pages up to depth four in their trees in order to exclude irrelevant entities, e.g., ``Lord of the Rings'' in DBpedia.
}

\subsection{Entity Linking Models}

% define problem.
% \paragraph{Problem Description}
Given a text span $s$ identified by the entity extraction model in \S\ref{extraction_model} (or with heuristics) and a reference KB, the goal of the entity linking model is to associate the span with the entity it refers to.
%$e_{ref} \in{} $\cal{K}. 
A span and its surrounding words are collectively referred to as a mention. 
We first identify a set of candidate entities that a given mention may refer to.
Then, we rank the candidate entities based on a score computed using a neural model trained on labeled data.

% The paragraph below can be removed if we want to save space
For example, given the string ``\ldots{} \textit{database of facts, an ILP system will} \ldots{}'', the entity extraction model identifies the span ``ILP'' as a possible entity and the entity linking model associates it with ``Inductive\_Logic\_Programming'' as the referent entity (from among other candidates like ``Integer\_Linear\_Programming'' or ``Instruction-level\_Parallelism'').

% describe model.
\paragraph{Datasets.} We used two datasets: i) a biomedical dataset formed by combining MSH \cite{jimeno:2011} and BC5CDR \cite{li:16} with UMLS as the reference KB, and ii) a CS dataset we curated using Wikipedia articles about CS concepts with DBpedia as the reference KB.

\paragraph{Candidate selection.} In a preprocessing step, we build an index which maps any token used in a labeled mention or an entity name in the KB to associated entity IDs, along with the frequency this token is associated with that entity. 
This is similar to the index used in previous entity linking systems \cite[e.g.,][]{bhagavatula:15} to estimate the probability that a given mention refers to an entity. 
At train and test time, we use this index to find candidate entities for a given mention by looking up the tokens in the mention. 
This method also serves as our baseline in Table \ref{tab:el_results} by selecting the entity with the highest frequency for a given mention.

\paragraph{Scoring candidates.} Given a mention (m) and a candidate entity (e), the neural model constructs a vector encoding of the mention and the entity. 
We encode the mention and entity using the functions $\mathbf{f}$ and $\mathbf{g}$, respectively, as follows:
\begin{align*}
\textbf{f}(\text{m}) &= [\textbf{v}_{\text{m.name}};\text{avg}(\textbf{v}_{\text{m.lc}}, \textbf{v}_{\text{m.rc}})], \nonumber \\
\textbf{g}(\text{e}) &= [\mathbf{v}_{\text{e.name}};\mathbf{v}_{\text{e.def}}],
\end{align*}
where m.surface, m.lc and m.rc are the mention's surface form, left and right contexts, and e.name and e.def are the candidate entity's name and definition, respectively.
$\mathbf{v}_\text{text}$ is a bag-of-words sum encoder for text.
We use the same encoder for the mention surface form and the candidate name, and another encoder for the mention contexts and entity definition.
%similar to the one used in \cite{chandrab:2018}. \wacomment{TODO: fix the citeomatic citation then add it.}

Additionally, we include numerical features to estimate the confidence of a candidate entity based on the statistics collected in the index described earlier. We compute two scores based on the word overlap of (i) mention's context and candidate's definition and (ii) mention's surface span and the candidate entity's name. Finally, we feed the concatenation of the cosine similarity between $\mathbf{f}(\text{m})$ and $\mathbf{g}(\text{e})$ and the intersection-based scores into an affine transformation followed by a sigmoid non-linearity to compute the final score for the pair (m, e).

\paragraph{Results.}
We use the Bag of Concepts F1 metric \cite{ling:15} for comparison.
Table \ref{tab:el_results} compares the performance of the most-frequent-entity baseline and our neural model described above.

%\wacomment{TODO: follow up with chandra to verify these results. I remember the difference being much smaller before.}
\begin{table}[t]
\centering
\begin{tabular}{l|c|c}
\toprule
  &  CS  & Bio \\
\midrule
Baseline &  84.2  &  54.2  \\
Neural  &  84.6  &  85.8\\
\bottomrule
\end{tabular}
\caption{The Bag of Concepts F1 score of the baseline and neural model on the two curated datasets.}
\label{tab:el_results}
\end{table}

\section{Other Research Problems}
\label{sec:author_disambiguation}
\label{sec:others}
In the previous sections, we discussed how we construct the main components of the literature graph.
In this section, we briefly describe several other related challenges we are actively working on.
%\lwcomment{This sentence is funny. Maybe just remove the last part: We briefly describe several other related challenges we are actively working on.}\wacomment{done}

\paragraph{Author disambiguation.}
Despite initiatives to have global author IDs ORCID and ResearcherID, most publishers provide author information as names (e.g., arXiv).
However, author names cannot be used as a unique identifier since several people often share the same name.
Moreover, different venues and sources use different conventions in reporting the author names, e.g., ``first initial, last name'' vs.~``last name, first name''.
Inspired by \newcite{culotta:07}, we train a supervised binary classifier for merging pairs of author instances and use it to incrementally create author clusters.
We only consider merging two author instances if they have the same last name and share the first initial.
If the first name is spelled out (rather than abbreviated) in both author instances, we also require that the first name matches.

\paragraph{Ontology matching.}
Popular concepts are often represented in multiple KBs.
For example, the concept of ``artificial neural networks'' is represented as entity ID D016571 in the MESH ontology, and represented as page ID `21523' in DBpedia.
Ontology matching is the problem of identifying semantically-equivalent entities across KBs or ontologies.\footnote{Variants of this problem are also known as deduplication or record linkage.}
%We address ontology matching as a binary classification problem given a pair of entities, using string-based heuristics to find candidate pairs.

\paragraph{Limited KB coverage.}
The convenience of grounding entities in a hand-curated KB comes at the cost of limited coverage.
Introduction of new concepts and relations in the scientific literature occurs at a faster pace than KB curation, resulting in a large gap in KB coverage of scientific concepts.
In order to close this gap, we need to develop models which can predict textual relations as well as detailed concept descriptions in scientific papers.
For the same reasons, we also need to augment the relations imported from the KB with relations extracted from text.
Our approach to address both entity and relation coverage is based on distant supervision \cite{mintz:09}.
In short, we train two models for identifying entity definitions and relations expressed in natural language in scientific documents, and automatically generate labeled data for training these models using known definitions and relations in the KB.
%automatically label textual descriptions in the paper text for known entities in the KB, then train a sequence tagger to distinguish the linguistic patterns used to introduce (new) concepts.

We note that the literature graph currently lacks coverage for important entity types (e.g., affiliations) and domains (e.g., physics).
Covering affiliations requires small modifications to the metadata extraction model followed by an algorithm for matching author names with their affiliations.
In order to cover additional scientific domains, more agreements need to be signed with publishers.

\paragraph{Figure and table extraction.}
Non-textual components such as charts, diagrams and tables provide key information in many scientific documents, but the lack of large labeled datasets has impeded the development of data-driven methods for scientific figure extraction.
In \newcite{siegel:18}, we induced high-quality training labels for the task of figure extraction in a large number of scientific documents, with no human intervention. 
To accomplish this we leveraged the auxiliary data provided in two large web collections of scientific documents (arXiv and PubMed) to locate figures and their associated captions in the rasterized PDF.
We use the resulting dataset to train a deep neural network for end-to-end figure detection, yielding a model that can be more easily extended to new domains compared to previous work.

\paragraph{Understanding and predicting citations.}
The citation edges in the literature graph provide a wealth of information (e.g., at what rate a paper is being cited and whether it is accelerating), and opens the door for further research to better understand and predict citations. 
For example, in order to allow users to better understand what impact a paper had and effectively navigate its citations, we experimented with methods for classifying a citation as important or incidental, as well as more fine-grained classes \cite{valenzuela:15}.
The citation information also enables us to develop models for estimating the potential of a paper or an author.
In \newcite{weihs:17}, we predict citation-based metrics such as an author's h-index and the citation rate of a paper in the future. 
Also related is the problem of predicting which papers should be cited in a given draft \cite{bhagavatula:18}, which can help improve the quality of a paper draft before it is submitted for peer review, or used to supplement the list of references after a paper is published.
%\lwcomment{might be worth mentioning alt-metrics}\wacomment{not entirely sure how to best talk about alt-metrics in this context. feels like a distraction.}

\section{Conclusion and Future Work}\label{sec:conclusion}
In this paper, we discuss the construction of a graph, providing a symbolic representation of the scientific literature.
We describe deployed models for identifying authors, references and entities in the paper text, and provide experimental results to evaluate the performance of each model.

Three research directions follow from this work and other similar projects, e.g., \newcite{hahnpowell:17,wu:14}:
i) improving quality and enriching content of the literature graph (e.g., ontology matching and knowledge base population).
ii) aggregating domain-specific extractions across many papers to enable a better understanding of the literature as a whole (e.g., identifying demographic biases in clinical trial participants and summarizing empirical results on important tasks).
iii) exploring the literature via natural language interfaces.

In order to help future research efforts, we make the following resources publicly available: 
metadata for over 20 million papers,\footnote{\url{http://labs.semanticscholar.org/corpus/}} 
meaningful citations dataset,\footnote{\url{http://allenai.org/data.html}}
models for figure and table extraction,\footnote{\url{https://github.com/allenai/deepfigures-open}}
models for predicting citations in a paper draft \footnote{\url{https://github.com/allenai/citeomatic}} and
models for extracting paper metadata,\footnote{\url{https://github.com/allenai/science-parse}}
among other resources.\footnote{\url{http://allenai.org/software/}}

\bibliography{review}
\bibliographystyle{acl_natbib}

%\clearpage
%\begin{appendices}
%%%%%%%%%%%%%%%%%%%%%%%%%%%%%%%%%%%%%%%%%%%%%%%%%%%%%%

%\section{Unnecessary details.}

%\end{appendices}
\end{document}